\documentclass[
]{ceurart}

\usepackage{soul} 
\usepackage{array}
\usepackage{tabularx} 
\usepackage{booktabs}
\usepackage{graphicx} 
\usepackage{wrapfig}
\usepackage{listings}
\usepackage{xcolor}
 \usepackage[normalem]{ulem}
\usepackage[utf8]{inputenc} 
\usepackage{tcolorbox}
\usepackage [english]{babel}
\usepackage [autostyle, english = american]{csquotes}
\MakeOuterQuote{"}

\usepackage{geometry}
\usepackage{caption}
\copyrightyear{2025}
\copyrightclause{Copyright for this paper by its authors.
  Use permitted under Creative Commons License Attribution 4.0
  International (CC BY 4.0).}

\conference{In: R. Campos, A. Jorge, A. Jatowt, S. Bhatia, M. Litvak (eds.): Proceedings of the Text2Story'25 Workshop, Lucca (Italy), 10-April-2025}

\title{Can Zero-Shot Commercial API's Deliver Regulatory-Grade Clinical Text De-Identification?}

\author[1]{Veysel Kocaman}[%
orcid=0000-0002-0065-6478,
email=veysel@johnsnowlabs.com]

\author[1]{Muhammed Santas}[%
orcid=0009-0009-9867-9624,
email=muhammed@johnsnowlabs.com]

\author[1]{Yigit Gul}[%
orcid=0009-0007-7184-9879,
email=yigit@johnsnowlabs.com]

\author[1]{Mehmet Butgul}[%
email=mehmet@johnsnowlabs.com]

\author[1]{David Talby}[%
email=david@johnsnowlabs.com]

\address[1]{John Snow Labs inc. 16192 Coastal Highway,
Lewes, DE 19958, USA}

\begin{document}
\maketitle
\begin{abstract}
We evaluate the performance of four leading solutions for de-identification of unstructured medical text - Azure Health Data Services, AWS Comprehend Medical, OpenAI GPT-4o, and John Snow Labs - on a ground truth dataset of 48 clinical documents annotated by medical experts. The analysis, conducted at both entity-level and token-level, suggests that John Snow Labs' Medical Language Models solution achieves the highest accuracy, with a 96\% F1-score in protected health information (PHI) detection, outperforming Azure (91\%), AWS (83\%), and GPT-4o (79\%). John Snow Labs is not only the only solution which achieves regulatory-grade accuracy (surpassing that of human experts) but is also the most cost-effective solution: It is over 80\% cheaper compared to Azure and GPT-4o, and is the only solution not priced by token. Its fixed-cost local deployment model avoids the escalating per-request fees of cloud-based services, making it a scalable and economical choice.

\end{abstract}


\section{Introduction}

Electronic Health Records (EHRs) are now widespread across the United States healthcare system, with adoption rates surpassing 96\% in acute care hospitals and 86\% among office-based physicians \cite{myrick2019percentage}. Although structured data, such as billing and claims information, constitutes a substantial component of EHRs, a significant proportion of clinical information remains in unstructured formats, including progress notes, discharge summaries, radiology reports, and pathology reports. This unstructured data contains valuable contextual details essential for comprehensive patient care. Its secondary use in research has gained increasing importance, with potential benefits in areas such as population health management, real-world evidence generation, patient safety enhancements, and drug discovery. However, processing unstructured data poses substantial ethical and technical challenges. The inherent variability of free-text documentation complicates efforts to preserve privacy, as sensitive patient information is frequently embedded within clinical narratives.

Given the highly sensitive nature of this data, it must undergo a de-identification process before use. De-identification involves removing or obscuring personal health information (PHI) from medical records to protect patient privacy. De-identified data refers to health information that has been stripped of all “direct identifiers”—elements that could uniquely identify an individual. The Health Insurance Portability and Accountability Act (HIPAA) Safe Harbor guidelines define 18 such direct identifiers (U.S. Department of Health \& Human Services, 2023 ) \cite{HHS2023}, though any additional data points capable of uniquely identifying a patient must also be considered. The federally regulated HIPAA Privacy Rule outlines two primary methods for de-identifying PHI: Expert Determination and Safe Harbor.

\begin{figure}[h]
    \centering
    \includegraphics[width=0.99\textwidth]{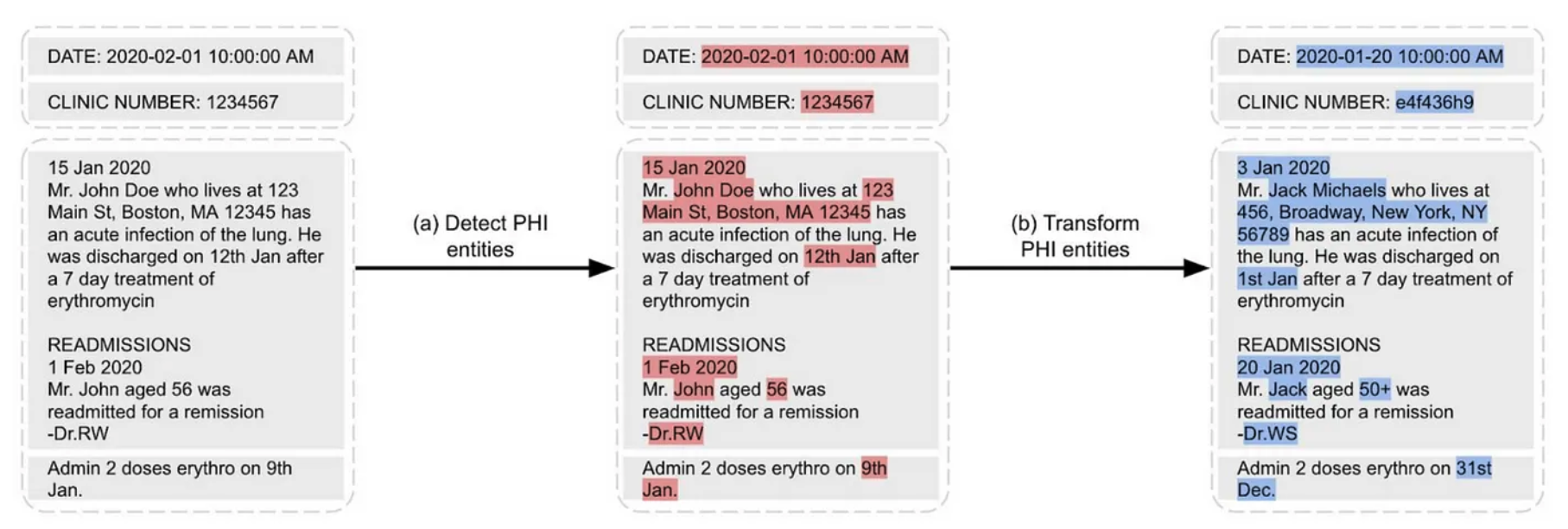}    
    \caption{De-Identification process identifies potential pieces of content with personal information about patients and removes them by replacing them with semantic tags or fake entities.}
    \label{fig:intro}
\end{figure}

Recent studies suggest that deep learning-based automated de-identification models can surpass human annotators in identifying PHI, with hybrid approaches demonstrating the greatest potential \cite{negash2023identification}. Once the de-identification criteria for a specific dataset have been established, advanced technologies can be employed to automate the detection of protected health information (PHI) in both structured and unstructured data. The combination of machine learning techniques and sophisticated Natural Language Processing (NLP) algorithms has markedly enhanced the capacity to identify and flag PHI across various data formats. To streamline the de-identification process, researchers can utilize Large Language Models (LLMs), specialized NLP models, and cloud provider APIs for processing extensive clinical datasets. However, the task of handling ambiguous or novel instances of identifiable information remains challenging, necessitating continuous improvement of these automated tools to strike a balance between efficiency and the nuanced interpretation required in healthcare settings. It is worth noting that while LLMs offer powerful capabilities, their application in de-identifying sensitive data (PHI) may be considered excessive or potentially unreliable for certain use cases, particularly when a high degree of customization is required. The choice of technology should be carefully evaluated based on the specific requirements of the de-identification task and the desired level of precision.

This study examines the performance and compares de-identification services, developed by us and named as Healthcare NLP library, AWS Comprehend Medical, and Azure Health Data Services, with a focus on their accuracy when applied to a dataset annotated by healthcare experts. The comparison of these services provides valuable insights into their respective strengths and limitations, enabling informed decision-making for researchers, developers, and organizations seeking appropriate de-identification tools. Additionally, this comprehensive analysis equips stakeholders with the necessary information to select the most suitable tool based on accuracy, compliance, cost-effectiveness, and scalability for processing sensitive healthcare data.

For researchers, this analysis helps identify the most accurate, reliable, and cost-effective service for processing sensitive data, which is crucial for maintaining data integrity in clinical studies. Developers benefit from understanding the ease of integration and API flexibility of each service, essential factors for building scalable solutions that can handle large volumes of clinical data \cite{kocaman2023rwd143}. Organizations, especially in the healthcare and finance sectors, gain valuable insights into the compliance capabilities and performance of these tools, ensuring that the chosen solution aligns with regulatory requirements while enhancing operational efficiency.

The comparison highlights variations in performance among the evaluated services. Our Healthcare NLP library achieved the highest accuracy, with macro and weighted average F1-scores of 96\% and 99\%, respectively, followed by Azure Health Data Services with 85\% macro and 99\% weighted average F1-scores, and AWS Comprehend Medical with 80\% macro and 98\% weighted average F1-score. However, performance may vary based on specific use cases and dataset characteristics. Additionally, a cost analysis for processing one million clinical notes (each containing 5,250 characters) revealed that the Healthcare NLP library is the most cost-effective option, followed by Azure Health Data Services and AWS Comprehend Medical.

\section{Background and Related Work}

The de-identification of unstructured data has been extensively studied, with various Natural Language Processing (NLP) approaches proposed over the years \cite{nadkarni2011natural, khin2018deep}. This process can be divided into two main subtasks: first, identifying Protected Health Information (PHI) within the text, and second, replacing those identifiers through either masking (substituting them with placeholder values) or obfuscation (replacing them with randomly generated values based on their type). Among these, the task of PHI identification has been the primary focus of research \cite{kocaman2023rwd143}.

Early de-identification systems in the clinical domain were predominantly rule-based, as seen in the work of Sweeney \cite{sweeney1996replacing} and Gupta et al. \cite{gupta2004evaluation}. These systems relied on regular expressions, syntactic rules, and specialized dictionaries to detect PHI in text. While rule-based approaches are effective in identifying structured PHI elements such as phone numbers, email addresses, and license numbers, they struggle with more complex entities, including personal names, professions, and hospital names \cite{liu2017identification}. Rule-based systems, while effective in specific contexts, often exhibit limited generalizability across diverse datasets. These systems typically require substantial modifications to their underlying dictionaries and rule sets when applied to new environments, hindering their adaptability and scalability in varied clinical settings.

The field of automated PHI detection and de-identification has seen significant advancements in recent years, with several major cloud providers and specialized services offering solutions to address the growing need for secure handling of sensitive healthcare data. The concept of automatic de-identification gained prominence in 2014 through the Informatics for Integrating Biology and the Bedside (i2b2) project, which introduced a pioneering academic NLP challenge focused on automatically detecting PHI identifiers from medical records \cite{uzuner2007evaluating}. This initiative accelerated research and development of Machine Learning and Deep Learning algorithms for robust PHI identification, laying the groundwork for more sophisticated approaches that are now being implemented by major cloud service providers.

Recent research suggests that deep learning-based automated de-identification models can surpass human annotators in PHI identification, with hybrid approaches demonstrating the greatest potential \cite{negash2023identification}. In the current landscape, several key players have emerged with offerings designed to streamline the process of PHI detection and de-identification. Several studies have conducted performance comparisons of PHI detection systems, providing valuable insights into the effectiveness of various de-identification approaches. These comparisons are crucial for researchers and healthcare organizations seeking to implement efficient and accurate de-identification processes while maintaining data utility for secondary use in research and analytics.

A notable study by Steinkamp et al. \cite{steinkamp2020evaluation} evaluated five publicly available de-identification tools on a large corpus of narrative-text radiology reports. The research assessed token-level recall, precision, and F1 scores for each tool across various PHI subcategories. The study found that machine learning systems outperformed rule-based systems, with the best-performing system (NeuroNER) achieving a token-level F1 score of 93.6\%. However, this performance was still below the acceptable level for clinical use (95\% recall) on sensitive categories of PHI.



Recent advancements in Large Language Models (LLMs) have prompted researchers to investigate their potential for de-identifying clinical notes. A study by Altalla et al. \cite{altalla2025evaluating} compared the de-identification performance of GPT-3.5 and GPT-4, revealing GPT-4's superior capabilities in this domain. The study, published on January 31, 2025, reported that GPT-4 achieved remarkable results with a precision of 0.9925, recall of 0.8318, F1 score of 0.8973, and accuracy of 0.9911, significantly outperforming its predecessor, GPT-3.5.

Despite these promising results, the application of LLMs for de-identification presents several challenges. The nascent stage of LLM utilization in this field raises concerns regarding the privacy and security of health data, particularly when employing API-based models \cite{liu2023deid}. Moreover, LLMs may encounter difficulties in striking a balance between effective de-identification and preserving the clinical utility of notes, potentially altering non-sensitive information crucial for research and analysis \cite{sarkar2024identification}. The variation in performance across different datasets highlights the need for continued development to achieve consistent and reliable results across diverse clinical settings.

This study aims to contribute to previous performance comparisons in PHI entity recognition and assist researchers and decision-makers in selecting the most suitable tool for processing large-scale datasets with high accuracy and cost-effectiveness. To achieve this, we compare three widely used and advanced de-identification tools that incorporate state-of-the-art models while ensuring consistency: Our Healthcare NLP library, Azure Health Data Services, AWS Comprehend Medical and GPT4o, a state-of-the-art commercial multi-modal LLM.

\section{Experiments and Results}

\subsection{The Deidentification Solutions}

In this section, we will provide brief information for each de-identification solution that supports different set of PHI entities. The list of PHI entities supported by each model is shared in Table \ref{tab:deid_tools}.

\subsubsection{John Snow Labs Healthcare NLP Library \& Medical Language Models}

The Healthcare NLP library by John Snow Labs is a powerful component of Spark NLP platform \cite{kocaman2021spark}, specifically designed to facilitate NLP tasks within the healthcare domain \cite{kocaman2022accurate}. This library offers over 2,500 pre-trained models and pipelines tailored for medical data, enabling accurate information extraction, named entity recognition (NER) for clinical and medical concepts, and robust text analysis capabilities. 
Regularly updated with advanced algorithms, it helps healthcare professionals derive meaningful insights from unstructured medical data sources such as electronic health records, clinical notes, and biomedical literature.

Additionally, the library features custom large language models (LLMs) in various sizes and quantization levels for tasks like medical note summarization, question answering, retrieval-augmented generation (RAG), and healthcare-related conversational interactions. It also provides a robust solution for de-identifying medical records using advanced NER models to automatically detect and remove PHI from clinical notes. This ensures compliance with privacy regulations while preserving data utility for research, enabling secure data sharing, enhancing patient privacy, and promoting innovation in medical research.

The Healthcare NLP library allows users to create custom de-identification pipelines targeting specific labels or to utilize pre-trained pipelines with two lines of code to de-identify a broad range of entities. These entities include AGE, CONTACT, DATE, ID, LOCATION, NAME, PROFESSION, CITY, COUNTRY, DOCTOR, HOSPITAL, IDNUM, MEDICALRECORD, ORGANIZATION, PATIENT, PHONE, STREET, USERNAME, ZIP, ACCOUNT, LICENSE, VIN, SSN, DLN, PLATE, IPADDR, EMAIL, and more. In \textit{Beyond Accuracy: Automated De-Identification of Large Real-World Clinical Text Datasets} [4], the de-identification process is explained in detail, describing the implementation of a hybrid context-based model architecture for automated clinical note processing.  

In this study, a pre-trained de-identification pipeline was utilized, specifically designed to extract and de-identify entities such as NAME, IDNUM, CONTACT, LOCATION, AGE, and DATE. Notably, this pipeline operates independently of any large language model (LLM) components.

\subsubsection{Azure Health Data Services}

Azure Health Data Services' de-identification service is designed to safeguard sensitive health information while maintaining data utility. This API employs advanced natural language processing techniques to identify, label, redact, or surrogate PHI in unstructured medical texts. The service provides three essential operations: Tag, Redact, and Surrogate, which allow healthcare organizations to process various types of clinical documents securely and efficiently. By utilizing machine learning algorithms, the service can detect HIPAA's 18 identifiers and other PHI entities, ensuring compliance with various regional privacy regulations such as GDPR and CCPA.


\subsubsection{Amazon Comprehend Medical}

Amazon Comprehend Medical is a HIPAA-eligible natural language processing (NLP) service that leverages machine learning to extract valuable health data from unstructured medical text. This tool quickly and accurately identifies medical entities such as conditions, medications, dosages, tests, treatments, and Protected Health Information (PHI) from various clinical documents including physician’s notes, discharge summaries, and test results. With its ability to understand context and relationships between extracted information, AWS Comprehend Medical offers a robust solution for healthcare professionals and researchers looking to automate data extraction, improve patient care, and streamline clinical workflows.

\subsubsection{Open AI GPT-4o for Deidentification}

GPT-4o is a multi-modal model that offers improvements in response times and classification accuracy compared to GPT-4, which could theoretically enhance the precision of identifying and redacting sensitive information via prompting. While GPT-3.5 and GPT-4 have been extensively studied for their de-identification capabilities, particularly in processing medical text, GPT-4o presents an intriguing option due to its enhanced performance over GPT-4 in various tasks. However, no formal study has yet evaluated GPT-4o’s de-identification capabilities. Given the importance of PHI redaction in healthcare AI applications, understanding the model's strengths and limitations in this area remains crucial. Despite these advantages, its effectiveness in de-identification remains speculative without empirical studies directly assessing its performance.
While there are cost-effective alternatives for de-identification, we opted for GPT-4o due to its widespread adoption, strong presence in research, and its demonstrated advancements over previous models.

\subsection{Dataset}
The annotation of patient identifiers within clinical data is a critical process in healthcare research and data management. This study employed a comprehensive annotation methodology utilizing the John Snow Labs' Annotation Lab software, which facilitated a multi-stage approach to entity recognition and labeling. The process began with a pre-annotation step using deep learning models to extract initial entities, followed by human refinement guided by a dynamic annotation guide. This iterative approach, involving multiple rounds of review and correction, ensured high accuracy and adaptability throughout the fine-tuning and evaluation phases \cite{xu2024accelerating}.

The dataset employed in this study comprised 48 clinical notes meticulously annotated by our domain experts. The dataset was specifically curated to facilitate the evaluation of de-identification systems in a healthcare context. Expert annotations focused on six key entity types: IDNUM, LOCATION, DATE, AGE, NAME, and CONTACT. These entities represent critical categories of Protected Health Information (PHI) that are commonly subject to de-identification under regulatory frameworks such as the Health Insurance Portability and Accountability Act (HIPAA) and the General Data Protection Regulation (GDPR).

The selection of these entity types was motivated by their frequent occurrence in clinical narratives and their significance in ensuring patient privacy. Identifiers such as patient names, contact details, and unique ID numbers pose a high risk of re-identification if not properly anonymized. Similarly, location information, age, and date-related details can contribute to indirect re-identification, necessitating robust de-identification strategies. By centering the benchmark on these entities, this study ensures that the performance evaluation remains directly aligned with real-world de-identification challenges in healthcare settings.

To enhance reproducibility, the benchmark dataset utilized in this study has been made publicly available in a dedicated repository\cite{deidentification_benchmark}. This ensures transparency and facilitates further research in the field of healthcare de-identification.

\subsection{Comparison of the Solutions}

The most significant difference between these tools lies in their adaptability. Azure Health Data Services, Amazon Comprehend Medical and GPT-4o are API-based, black-box cloud solutions, making modifying or adapting results to specific needs impossible. On the other hand, the Healthcare NLP library’s de-identification pipeline can be loaded and utilized with just two lines of code. The pipeline outputs can be customized by adjusting its stages to meet specific needs, and it can also be used locally with no internet connection.

\subsubsection{Evaluation Criteria}

In this benchmark study, we employed two distinct approaches to compare accuracy:

\subsubsection{Entity-Level Evaluation}
Since de-identifying PHI data is a critical task, we evaluated how well de-identification tools detected entities present in the annotated dataset, regardless of their specific labels in the ground truth. The detection outcomes were categorized as:

\begin{itemize}
  \item \textbf{full\_match:} The entire entity was correctly detected.
  \item \textbf{partial\_match:} Only a portion of the entity was detected.
  \item \textbf{not\_matched:} The entity was not detected at all.
\end{itemize}

For example, for the text: "Patient John Doe was admitted to Boston General Hospital on 01/12/2023.", the ground truth entity "John Doe (NAME)" could have the following predicted entities:

\begin{itemize}
   \item Predicted Entity: "John Doe (NAME)" ==> \textbf{full\_match}
   \item Predicted Entity: "John" ==> \textbf{partial\_match}
   \item Predicted Entity: "Patient" ==> \textbf{not\_matched}
 \end{itemize}

For evaluation results, please refer to Figure A1 in the Appendix section.

\subsubsection{Token-Level Accuracy}
The text in the annotated dataset was tokenized, and the ground truth labels assigned to each token were compared with predictions made by the Healthcare NLP library, Amazon Comprehend Medical, Azure Health Data Services, and GPT-4o model. Classification reports were generated for each tool, comparing their precision, recall, and F1 scores.
Token-level evaluation results are presented in Figure A2 in the Appendix section.

\subsection{Methodology}

In this study, differences were observed between the predictions generated by the de-identification services and the ground truth annotations. The ground truth dataset utilized generic entity labels; for instance, all names were annotated as \textbf{NAME}, rather than distinguishing between \textbf{PATIENT\_NAME} and \textbf{DOCTOR\_NAME}. To ensure consistency in evaluation, the predicted labels from the de-identification tools were mapped to their corresponding ground truth labels.

To maintain a fair comparison, entities that did not have a direct mapping to the ground truth labels—such as \textbf{PROFESSION}, \textbf{ORGANIZATION}, and other non-essential entity types—were excluded from the predictions before conducting the performance evaluation. This preprocessing step ensured that the assessment focused solely on the six critical entity types relevant to healthcare de-identification. Entity mapping table showing entity mapping across different providers can be seen at Table \ref{app:tab:deid_entities}. After obtaining the model predictions and applying the preprocessing steps, the entity distribution was summarized in Table \ref{tab:chunk_label_counts}.
 While evaluating GPT4o, we used a one-shot prompt to provide the model some sample PHI entity extraction tasks (the prompt is shared in the Appendix).The model was configured with a temperature of 1 and executed as a single run, while all other parameters were maintained at their default settings to ensure consistency in evaluation.

\section{Experiments and Results}

\subsection{Performance Evaluation}

The final results can be found at Table \ref{tab:phi_nlp_comparison}. The entity-level and token-level evaluations including comparative analyses and benchmark scores can be found in the Appendix. 

\begin{table}[h]
    \centering
    \caption{Healthcare NLP, Azure, Amazon, and GPT-4o PHI Recognition and Benchmark Comparison (Sample size: 45172 PHI entities).}
    \small
    \resizebox{\textwidth}{!}{
    \begin{tabular}{lccc|ccc|ccc|ccc}
        \toprule
        \textbf{Metric / Entity} & \multicolumn{3}{c}{\textbf{Healthcare NLP}} & \multicolumn{3}{c}{\textbf{Azure}} & \multicolumn{3}{c}{\textbf{Amazon}} & \multicolumn{3}{c}{\textbf{GPT-4o}} \\
        \cmidrule(lr){2-4} \cmidrule(lr){5-7} \cmidrule(lr){8-10} \cmidrule(lr){11-13}
        & Precision & Recall & F1-score & Precision & Recall & F1-score & Precision & Recall & F1-score & Precision & Recall & F1-score \\
        
        \midrule
        AGE       & 0.96 & 1.00 & \textbf{0.98} & 0.94 & 0.45 & 0.61 & 1.00 & 0.41 & 0.58 & 0.87 & 0.50 & 0.64 \\
        CONTACT   & 0.96 & 0.97 & \textbf{0.97} & 0.73 & 0.88 & 0.80 & 0.78 & 0.72 & 0.75 & 0.67 & 0.53 & 0.59 \\
        DATE      & 0.97 & 0.99 & \textbf{0.98} & 0.91 & 0.99 & 0.95 & 0.90 & 0.97 & 0.93 & 0.79 & 0.72 & 0.75 \\
        IDNUM     & 0.98 & 0.94 & \textbf{0.96} & 0.78 & 0.93 & 0.85 & 0.95 & 0.86 & 0.91 & 0.70 & 0.92 & 0.80 \\
        LOCATION  & 0.93 & 0.92 & \textbf{0.93} & 0.89 & 0.87 & 0.88 & 0.52 & 0.74 & 0.61 & 0.82 & 0.72 & 0.76 \\
        NAME      & 0.92 & 0.94 & \textbf{0.93} & 0.92 & 0.89 & 0.90 & 0.85 & 0.76 & 0.80 & 0.79 & 0.82 & 0.80 \\
        O         & 1.00 & 1.00 & \textbf{1.00} & 1.00 & 0.99 & 0.99 & 0.99 & 0.99 & 0.99 & 0.99 & 0.99 & 0.99 \\    
        \midrule
        Macro Avg & 0.96 & 0.97 & \textbf{0.96} & 0.88 & 0.86 & 0.85 & 0.86 & 0.78 & 0.80 & 0.80 & 0.74 & 0.76 \\
        \specialrule{1.8pt}{1pt}{1pt}
        Non-PHI  & 1.00 & 1.00 & \textbf{1.00} & 1.00 & 0.99 & 0.99 & 0.99 & 0.99 & 0.99 & 0.99 & 0.99 & 0.99 \\
        PHI      & 0.96 & 0.97 & \textbf{0.96} & 0.91 & 0.92 & 0.91 & 0.81 & 0.85 & 0.83 & 0.81 & 0.77 & 0.79 \\
        \midrule
        Macro Avg & 0.98 & 0.98 & \textbf{0.98} & 0.95 & 0.96 & 0.95 & 0.90 & 0.92 & 0.91 & 0.90 & 0.88 & 0.89 \\
        \specialrule{1.8pt}{1pt}{1pt}
        \textbf{cost per 1M doc} & \multicolumn{3}{c}{\textbf{\$2,418}} & \multicolumn{3}{c}{\$13,125} & \multicolumn{3}{c}{\$14,525} & \multicolumn{3}{c}{\$21,400}\\
        \bottomrule
    \end{tabular}
    }
    \label{tab:phi_nlp_comparison}
\end{table}

\begin{figure}[h]
    \centering
    \includegraphics[width=0.99\textwidth]{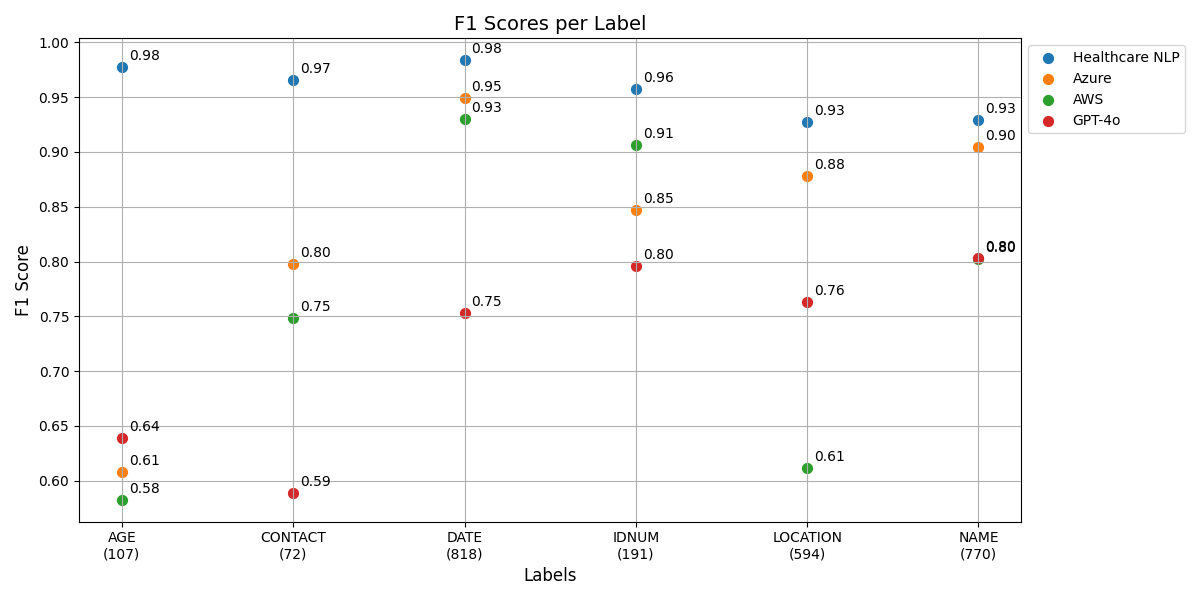}
    \caption{Visualization of the F1-Scores for each label}
    \label{fig:f1_score_per_label}
\end{figure}

The primary objective of de-identification is to accurately detect PHI entities. In this regard, we also wanted to evaluate binary classification performance  in which entities were classified as either PHI or non-PHI, disregarding specific subcategories. The PHI entity detection results are also summarized in Table~\ref{tab:phi_nlp_comparison} and Figure \ref{fig:f1_score_per_label}.

\subsection{Cost Estimation for De-identifying Clinical Data}

Cost is a critical factor when processing large-scale clinical datasets. To estimate expenses, we simulated the cost of de-identifying 1 million unstructured clinical notes, each averaging 5,250 characters.

The pricing estimates are as follows:
\begin{itemize}
    \item \textbf{Amazon Comprehend Medical:} Processing 1M documents costs approximately \textbf{\$14,525}.
    \item \textbf{Azure Health Data Services:} Processing 1M documents costs approximately \textbf{\$13,125}.
    \item \textbf{Open AI GPT-4o:} Processing 1M documents costs approximately \textbf{\$21,400}.
    \item \textbf{Healthcare NLP:} Using John Snow Labs' Healthcare NLP Prepaid on an EC2 c6a.8xlarge instance (\$1.2/hour), de-identifying PHI from 48 documents took 39.4 seconds. Extrapolating, processing 1M documents would take approximately 228 hours (9.5 days), but with proper scaling, it could be completed in a single day. The total estimated cost:
    \begin{itemize}
        \item \textbf{Infrastructure:} \$273
        \item \textbf{License:} \$2,145 (if one-month license cost set to \$7,000)
        \item \textbf{Total: \$2,418}
    \end{itemize}
\end{itemize}


\section{Conclusion}
In this study, we conducted a comparative analysis of the performance of Healthcare NLP, Amazon Comprehend Medical, Azure Health Data Services, and Open AI GPT-4o model on a ground truth dataset annotated by medical experts. The evaluation was performed at two levels: entity-level and token-level.

The entity-level analysis demonstrated that Healthcare NLP outperformed its counterparts in accurately capturing entities while minimizing missed detections. Azure Health Data Services exhibited the second-best performance, followed by Amazon Comprehend Medical. The GPT-4o model ranked fourth in this comparative assessment.

The token-level evaluation further reinforced these findings, with Healthcare NLP achieving the highest precision, recall, and F1-score. Azure Health Data Services, Amazon Comprehend Medical and GPT-4o followed in that order, indicating a consistent pattern of superior performance for Healthcare NLP across both evaluation metrics.

A key differentiator among these tools is their adaptability. While Azure Health Data Services, Amazon Comprehend Medical and GPT-4o function as API-based, black-box cloud solutions with no customization capabilities, Healthcare NLP provides a flexible and transparent framework. Its de-identification pipeline can be implemented with minimal coding effort, and users can modify pipeline stages to tailor the output to their specific requirements.

From a cost-effectiveness perspective, Healthcare NLP emerges as the most viable solution for large-scale clinical data processing. Unlike cloud-based services, which impose per-request pricing that escalates with increasing data volumes, Healthcare NLP allows for fixed-cost, local deployment. Even when processing substantial datasets, such as one billion clinical notes, its pricing remains stable over the same time period, providing a significant economic advantage over API-based alternatives.

In summary, Healthcare NLP consistently outperformed Azure Health Data Services, Amazon Comprehend Medical, and GPT-4o across all evaluation metrics by 5-10\%, achieving the highest accuracy while minimizing missed detections. Beyond its superior performance, its adaptability offers a crucial advantage over the black-box nature of cloud solutions, enabling users to customize de-identification pipelines to meet specific needs. Furthermore, its cost-effective deployment model presents substantial savings, making it a compelling alternative to API-based solutions.

\clearpage



\bibliography{references}

\appendix
\renewcommand{\thetable}{A\arabic{table}} 
\renewcommand{\thefigure}{A\arabic{figure}}
\setcounter{table}{0} 
\setcounter{figure}{0} 

\section*{Appendix}


\begin{table}[h]
    \centering
    \caption{Example of Original and NER Detection Text
    }
    \small
    \renewcommand{\arraystretch}{1.2}
    \begin{tabularx}{\textwidth}{lX X}
        \toprule
        \textbf{Type} & \textbf{Description} & \textbf{Example Text} \\
        \midrule
        Original & The original text with identifiable information & He is a \hl{60-year-old} male. \\
        \midrule
        NER Detection and Masking & Text with Named Entity Recognition (NER) applied. & He is a \hl{<AGE>} male. \\
        \bottomrule
    \end{tabularx}
    \label{tab:ner_example}
\end{table}

\begin{table}[h]
    \centering
    \caption{Comparison of De-identification Solutions}
    \small
    \renewcommand{\arraystretch}{1.2}
    \begin{tabular}{p{2cm} p{5cm} p{6cm}}
        \toprule
        \textbf{Tool} & \textbf{Entities De-identified} & \textbf{Key Features} \\
        \midrule
        Healthcare NLP Library & AGE, CONTACT, DATE, ID, LOCATION, NAME, PROFESSION, CITY, COUNTRY, DOCTOR, HOSPITAL, IDNUM, MEDICALRECORD, ORGANIZATION, PATIENT, PHONE, STREET, USERNAME, ZIP, ACCOUNT, LICENSE, VIN, SSN, DLN, PLATE, IPADDR, EMAIL & Highly flexible; the de-identification pipeline can be easily loaded with two lines of code and customized to meet specific requirements. Additionally, it can be used locally. \\
        \midrule
        Azure Health Data Services & DATE, DOCTOR, HOSPITAL, IDNUM, PATIENT, MEDICALRECORD, PHONE, AGE, STREET, STATE, CITY, HEALTHPLAN, PROFESSION, ZIP, EMAIL, ORGANIZATION, USERNAME, FAX, URL, LOCATIONOTHER, ACCOUNT, COUNTRYORREGION, SOCIALSECURITY & API-based, black-box solution; no direct control over results; suitable for integrated, cloud-based environments but lacks flexibility for task-specific adjustments. \\
        \midrule
        AWS Comprehend Medical & DATE, NAME, ADDRESS, ID, AGE, PHONE\_OR\_FAX, PROFESSION, URL, EMAIL & API-based, black-box solution; de-identification is limited to specific pre-configured models; lacks customization and flexibility for adapting results to specific needs. \\
         \midrule
        GPT-4o & No pre-built set of entities & API-based, black-box solution; de-identification is run via prompting.
        \\
        \bottomrule
    \end{tabular}
    \label{tab:deid_tools}
\end{table}

\clearpage
\subsection*{Evaluation Results}
The results obtained by comparing the predictions made by Healthcare NLP, AWS Comprehend Medical, and Azure Health Data Services with the ground truth entities are presented below.

\begin{figure}[h]
    \centering
    \includegraphics[width=0.8\textwidth]{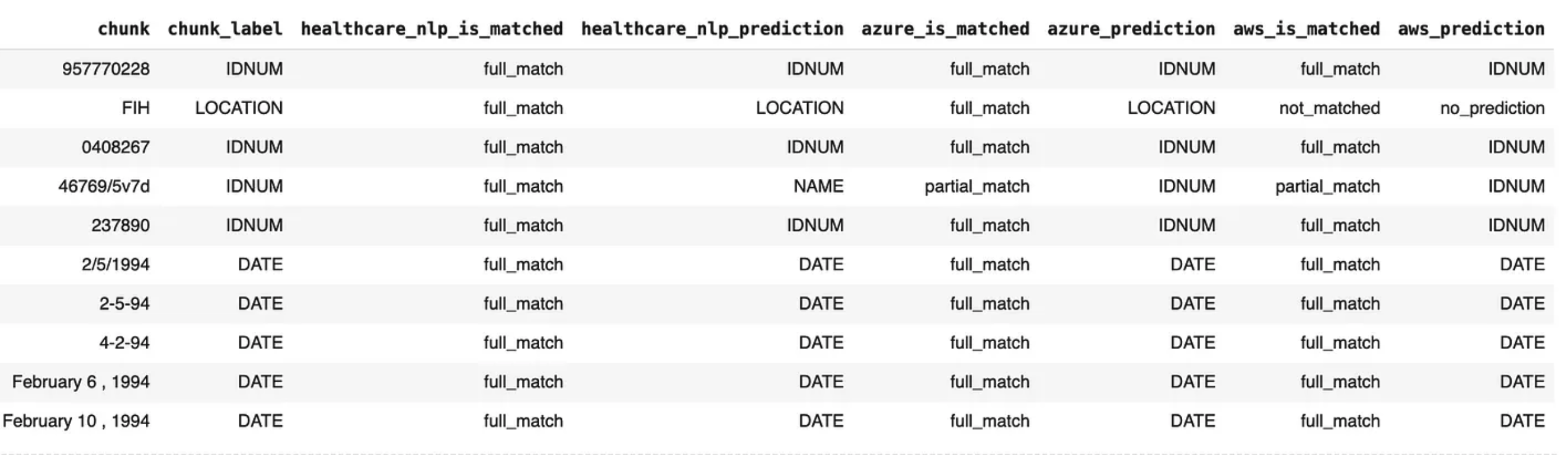}
    \caption{Entity Level Evaluation}
    \label{fig:entity_level_eval}
\end{figure}

To further analyze the performance of each de-identification tool, a token-level evaluation was conducted. This involved tokenizing the ground truth text and associating each token with the corresponding predicted labels from Healthcare NLP, Amazon Comprehend Medical, Azure Health Data Services and GPT-4o.

\begin{figure}[h]
    \centering
    \includegraphics[width=0.8\textwidth]{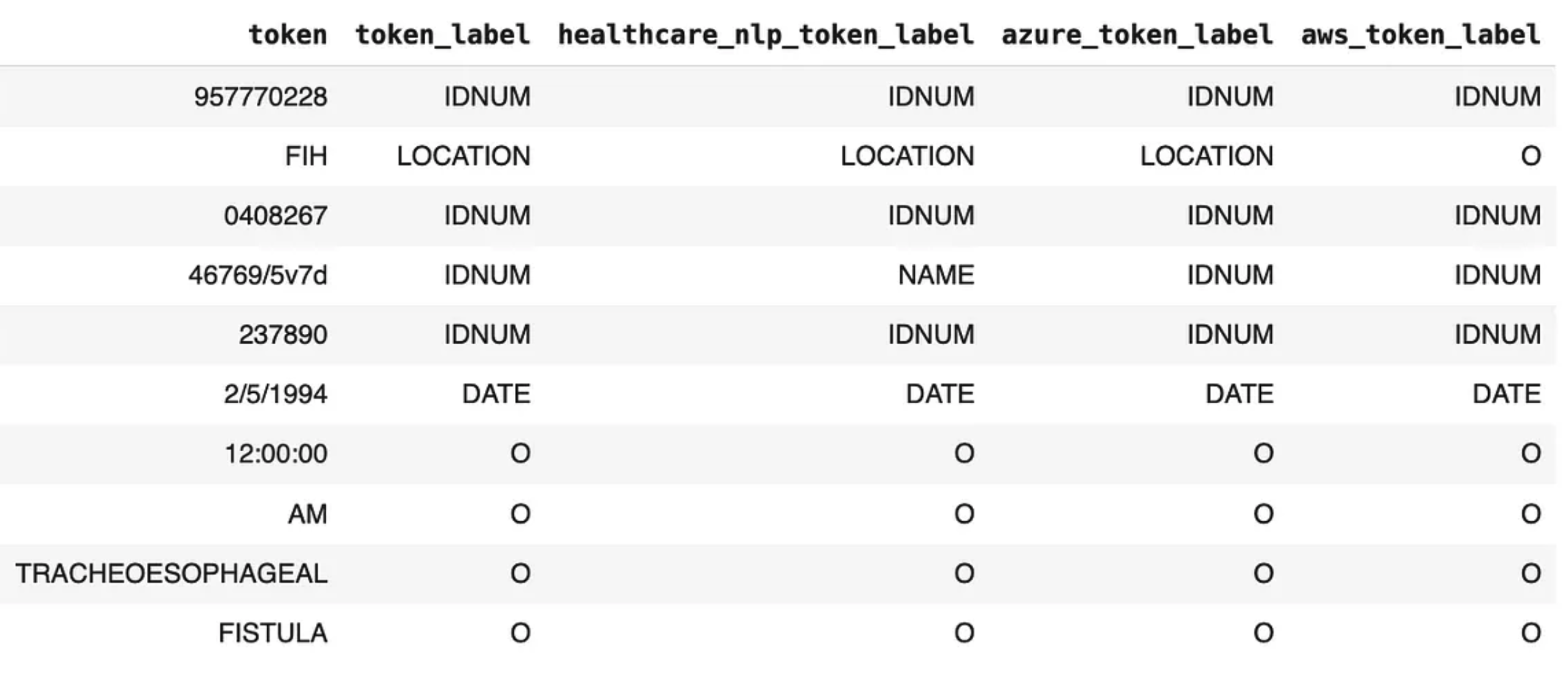}
    \caption{Token Level Evaluation}
    \label{fig:token_level_eval}
\end{figure}

\begin{table}[h]
    \centering
    \caption{Comparison of De-identified Entities}
    \small
    \renewcommand{\arraystretch}{1.5}
    \begin{tabular}{p{2cm}|p{3cm}|p{3cm}|p{3cm}|p{2cm}}
        \hline
        \textbf{Ground Truth Label} & \textbf{Healthcare NLP Library} & \textbf{Azure Health Data Services} & \textbf{AWS Medical Comprehend} & \textbf{GPT-4o} \\
        \midrule
        AGE & AGE & AGE & AGE & AGE \\
        \hline
        DATE & DATE & DATE & DATE & DATE \\
        \hline
        LOCATION & LOCATION, CITY, COUNTRY, HOSPITAL, STREET, ZIP & HOSPITAL, STREET, STATE, CITY, ZIP, LOCATIONOTHER, COUNTRYORREGION & ADDRESS & LOCATION \\
        \hline
        NAME & NAME, DOCTOR, PATIENT & DOCTOR, PATIENT & NAME & NAME \\
        \hline
        IDNUM & IDNUM, MEDICALRECORD, VIN, SSN, DLN, PLATE, ACCOUNT, LICENSE & IDNUM, MEDICALRECORD, ACCOUNT, SOCIALSECURITY & IDNUM & IDNUM \\
        \hline
        CONTACT & CONTACT, PHONE, EMAIL & PHONE, EMAIL, FAX & PHONE\_OR\_FAX, EMAIL & CONTACT \\
        \hline
    \end{tabular}
    
    \label{app:tab:deid_entities}
\end{table}

\begin{figure}[h]
    \centering
    \includegraphics[width=1.0\textwidth]{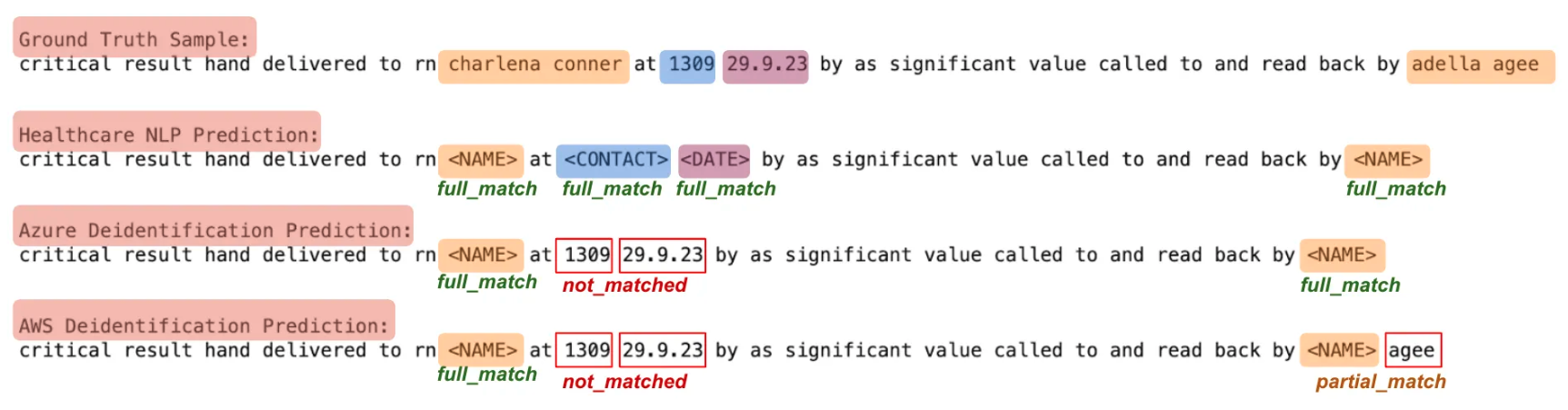}
    \caption{De-identification Results of the Tools on a Sample Text}
    \label{fig:deid-res}
\end{figure}

\begin{table}[h!]
\centering
\caption{Match Statistics for Healthcare NLP, Azure, AWS, and GPT-4o Predictions. The table shows the number of matches and their corresponding percentages for the different prediction models.}
\begin{tabular}{c|c|c|c|c}
\hline
\textbf{Match Type} & \textbf{Healthcare NLP} & \textbf{Azure} & \textbf{AWS} & \textbf{GPT-4o} \\
\hline
\text{Full Match}    & 1342 \, (90.7\%) & 1258 \, (85.0\%) & 1108 \, (74.9\%) & 983 \, (66.5\%) \\
\text{Partial Match} & 124 \, (8.4\%)  & 164 \, (11.1\%)  & 219 \, (14.8\%)  & 280 \, (18.9\%) \\
\text{Not Matched}   & 13 \, (0.9\%)   & 57 \, (3.8\%)   & 152 \, (10.3\%)  & 216 \, (14.6\%) \\
\hline
\end{tabular}
\end{table}

\begin{figure}[h!]
\begin{tcolorbox}[colframe=blue!50!black, colback=blue!10!white, coltitle=black, title=GPT-4o Prompt]
\footnotesize
You are an expert medical annotator with extensive experience in labeling medical entities within clinical texts. Your role is to accurately identify and annotate Protected Health Information (PHI) entities in the provided text, following the specified entity types.\\

\#\#\# Instructions:
\begin{itemize}
    \item[1] \textbf{Review the Text}: Carefully read the text to understand its medical context.
    \item[2] \textbf{Identify PHI Entities}: Locate any terms or phrases that represent PHI, based on the following entity types:
    \begin{itemize}
        \item IDNUM, LOCATION, DATE, AGE, NAME, CONTACT
    \end{itemize}
    \item[3] \textbf{Annotate Entities}: For each identified PHI, provide the start and end character indices, the entity type, and the exact text (chunk) of the entity.
    \item[4] \textbf{Response Format}: Return the annotations in a structured JSON format, as demonstrated in the examples below.
\end{itemize}

\#\#\# Example:

\textbf{Input Sentence:}\\
"MD Connect Call 11:59pm 2/16/69 from Dr. Hale at Senior Care Clinic Queen Creek, SD regarding Terri Bird."

\textbf{Annotated Entities:}\\

[

\{\{'begin': 24, 'end': 30, 'entity\_type': 'DATE', 'chunk': '2/16/69'\}\} \\
\{\{'begin': 42, 'end': 45, 'entity\_type': 'NAME', 'chunk': 'Hale'\}\} \\
\{\{'begin': 50, 'end': 67, 'entity\_type': 'LOCATION', 'chunk': 'Senior Care Clinic'\}\} \\
\{\{'begin': 69, 'end': 79, 'entity\_type': 'LOCATION', 'chunk': 'Queen Creek'\}\} \\
\{\{'begin': 83, 'end': 84, 'entity\_type': 'LOCATION', 'chunk': 'SD'\}\} \\
\{\{'begin': 96, 'end': 105, 'entity\_type': 'NAME', 'chunk': 'Terri Bird'\}\}

]

---

\textbf{Task:}\\
Extract all PHI entities from the text below. The entity types to identify are: \textbf{IDNUM, LOCATION, DATE, AGE, NAME, CONTACT}.

\textbf{Expected Output Format:}\\
\{ \texttt{entities:[}\\
    \texttt{\{'begin': <start\_index>, 'end': <end\_index>, 'entity\_type': '<entity\_type>', 'chunk': '<extracted\_text>'\}}\\
\texttt{]} \}

---

\textbf{Text to Annotate:}

\{text\}

---

\textbf{Your Response:}
\end{tcolorbox}
\label{fig:gpt_prompt}
\caption{Example of GPT-4o prompt for detecting Protected Health Information entities in clinical text}
\end{figure}

\begin{table}[h]
    \centering
    \caption{De-identified Chunk Label Counts for Different Tools}
    \footnotesize  
    \begin{tabular}{l|c|c|c|c|c}
        \hline
        \textbf{Chunk Label} & \textbf{Ground Truth} & \textbf{Healthcare NLP} & \textbf{Azure} & \textbf{AWS} & \textbf{GPT-4o} \\
        \hline
        DATE     & 582 & 591 & 617 & 571 & 566 \\
        NAME     & 380 & 401 & 393 & 333 & 391 \\
        LOCATION & 236 & 253 & 253 & 310 & 183 \\
        IDNUM    & 185 & 178 & 231 & 175 & 234 \\
        CONTACT  & 49  & 51  & 63  & 42  & 51  \\
        AGE      & 47  & 52  & 49  & 44  & 63  \\
        \hline
    \end{tabular}
    \label{tab:chunk_label_counts}
\end{table}

\lstdefinestyle{mystyle}{
    backgroundcolor=\color{gray!10},  
    basicstyle=\ttfamily\footnotesize, 
    breaklines=true,  
    frame=single, 
    captionpos=b, 
    numbers=left, 
    numberstyle=\tiny, 
    keywordstyle=\color{blue}, 
    commentstyle=\color{gray}, 
    stringstyle=\color{red}, 
}

\clearpage


    

\end{document}